\documentclass[10pt,twocolumn,letterpaper]{article}

\usepackage{cvpr}

\definecolor{cvprblue}{rgb}{0.21,0.49,0.74}
\usepackage[pagebackref,breaklinks,colorlinks,allcolors=cvprblue]{hyperref}

\def\paperID{32654} 
\def\confName{CVPR}
\def\confYear{2026}

\title{Active Inference for Micro-Gesture Recognition: EFE-Guided Temporal Sampling and Adaptive Learning}

\author{Weijia Feng$^1$, Jingyu Yang$^1$, Ruojia Zhang$^1$, Fengtao Sun$^1$, Qian Gao$^1$, Chenyang Wang$^{2}$\thanks{Corresponding author.}, \\
Tongtong Su$^1$, Jia Guo$^1$, Xiaobai Li$^{3,4}$, Minglai Shao$^5$\\
$^1$Tianjin Normal University $^2$Shenzhen University\\
$^3$ The State Key Laboratory of Blockchain and Data Security, Zhejiang University\\
$^4$Hangzhou High-Tech Zone (Binjiang) Institute of Blockchain and Data Security\\
$^5$School of New Media and Communications, Tianjin University\\
\tt\small $\{$weijiafeng, c04s316$\}$@tjnu.edu.cn, $\{$2411090052, 2410090022, 2311090028$\}$@stu.tjnu.edu.cn,
\\ \tt\small zrj20001127@163.com, chenyangwang@ieee.org, xiaobai.li@zju.edu.cn, shaoml@tju.edu.cn
}

\begin{document}
\maketitle
\begin{abstract}
Micro-gestures are subtle and transient movements triggered by unconscious neural and emotional activities, holding great potential for human–computer interaction and clinical monitoring. However, their low amplitude, short duration, and strong inter-subject variability make existing deep models prone to degradation under low-sample, noisy, and cross-subject conditions. This paper presents an active inference–based framework for micro-gesture recognition, featuring Expected Free Energy (EFE)-guided temporal sampling and uncertainty-aware adaptive learning. The model actively selects the most discriminative temporal segments under EFE guidance, enabling dynamic observation and information gain maximization. Meanwhile, sample weighting driven by predictive uncertainty mitigates the effects of label noise and distribution shift. Experiments on the SMG dataset demonstrate the effectiveness of the proposed method, achieving consistent improvements across multiple mainstream backbones. Ablation studies confirm that both the EFE-guided observation and the adaptive learning mechanism are crucial to the performance gains. This work offers an interpretable and scalable paradigm for temporal behavior modeling under low-resource and noisy conditions, with broad applicability to wearable sensing, HCI, and clinical emotion monitoring.
\end{abstract}    
\section{Introduction}
\label{sec:intro}

In recent years, with the continuous advancement of Human–Computer Interaction (HCI) technologies, systems capable of perceiving and interpreting subtle and unconscious human signals have been widely applied in areas such as public safety and affective computing~\cite{wang2013dense}. Among these signals, micro-gestures, which are involuntary, low-amplitude, and short-duration hand movements, have attracted increasing attention for their unique ability to reveal latent emotional states, psychological conditions, and hidden intentions~\cite{guo2024benchmarking}. Micro-gestures typically occur when facial expressions are suppressed or deliberately concealed. Due to their high authenticity and discriminability, they are difficult to imitate and have become valuable behavioral cues for affective computing, remote psychological assessment, and security monitoring~\cite{cormier2024enhancing}. Despite its potential, Micro-Gesture Recognition (MGR) remains a challenging task. Existing approaches are limited by the scarcity and inconsistency of annotated datasets, the inherently subtle and noise-sensitive nature of micro-gesture signals, and significant inter-subject variability. These factors substantially restrict the generalization capability of current models. 

Recent advances in deep learning have made notable progress in this field. Convolutional Neural Networks (CNNs)~\cite{ketkar2021convolutional} can extract hierarchical spatial representations from micro-gesture videos, while Recurrent Neural Networks (RNNs)~\cite{medsker2001recurrent} effectively model temporal dependencies and dynamic transitions, capturing the onset, duration, and offset of gestures. Spatio-temporal Transformers~\cite{tran2018closer,bertasius2021space,wang2025resource} further enhance performance by using self-attention to jointly process spatial and temporal information and adaptively focusing on informative regions and time steps. Although these approaches perform well in standard action recognition, they exhibit fundamental limitations in micro-gesture analysis. \textbf{\textit{i)}}, existing models passively process all spatio-temporal information, which makes them less sensitive to the transient and localized nature of micro-gestures. \textbf{\textit{ii)}}, they lack predictive uncertainty awareness, often showing overconfidence when facing ambiguous or low-quality samples, which leads to unreliable and unstable.

\begin{figure}[!t]
    \centering
    \includegraphics[width=0.9\linewidth]{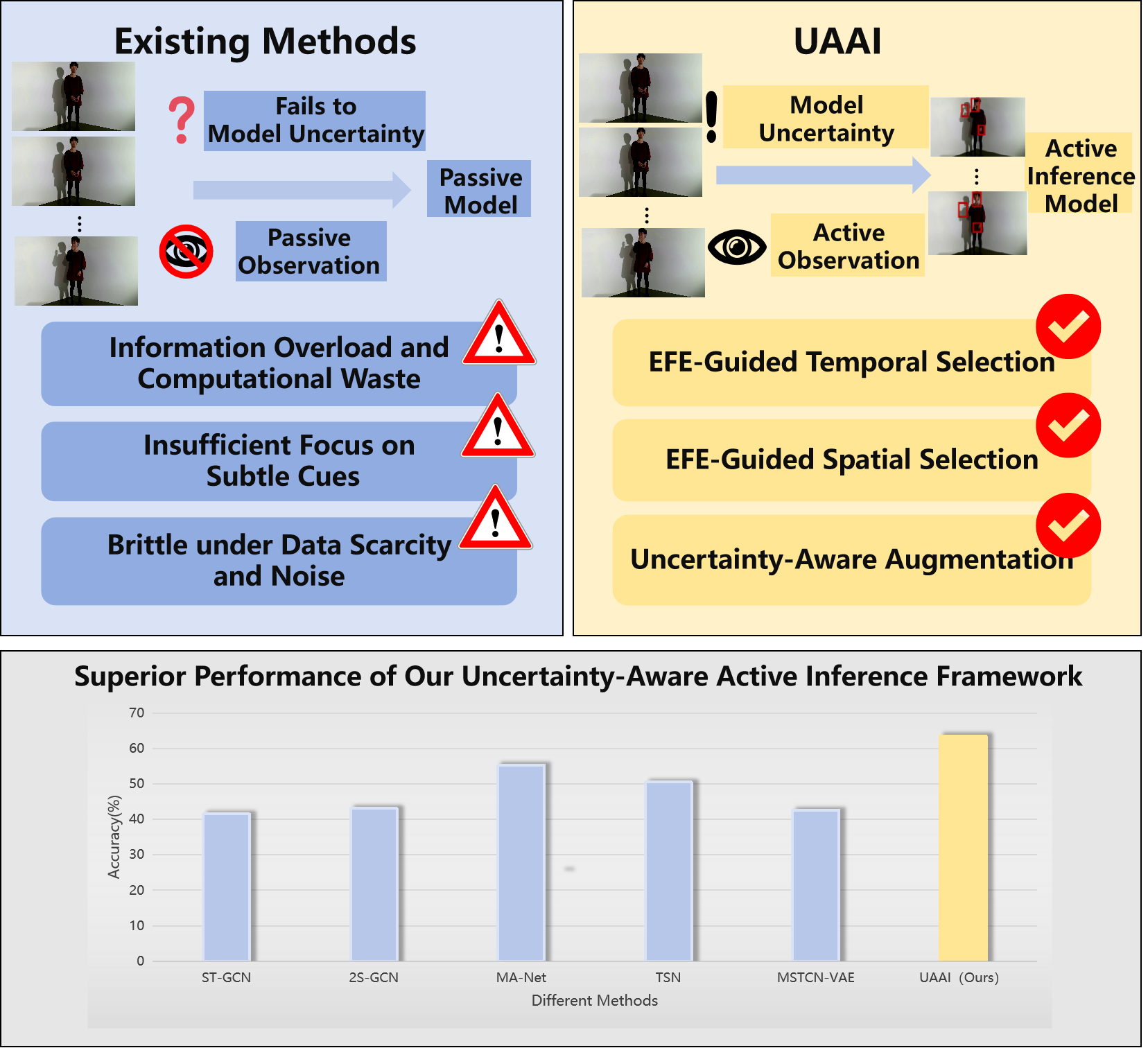}
    \caption{Overview of existing methods and UAAI}
    \vspace{-0.4cm}
    \label{fig: Overview}
\end{figure}
Recent works have explored dynamic keyframe selection for long-video understanding, such as Logic-in-Frames~\cite{guo2025logic} and VideoTree~\cite{wang2025videotree}, where frame selection aims to preserve semantic completeness for downstream reasoning. However, micro-gesture recognition differs fundamentally from long-video analysis. Micro-gestures are temporally sparse, fine-grained, and often lack explicit semantic structure. The core challenge lies in reducing predictive uncertainty rather than maintaining semantic coverage. Therefore, keyframe selection strategies designed for long-form semantic reasoning may not be directly suitable for micro-gesture recognition.

To overcome these limitations, we propose UAAI, an Uncertainty-Aware Active Inference framework for micro-gesture recognition, as shown in \Cref{fig: Overview}. Inspired by the principle of active inference, where an intelligent agent minimizes variational free energy to optimize perception and action, the proposed framework actively selects the most informative temporal frames and spatial regions. This mechanism reduces redundant computation and perceptual noise and improves recognition accuracy and robustness. UAAI applies the principle of variational free-energy minimization by jointly optimizing the accuracy and complexity terms, thereby enhancing both the perception and decision-making components of the model. During training, an active observation policy dynamically selects key frames and focuses on the most salient regions, while an uncertainty-aware augmentation (UMIX) mechanism adjusts sample weights based on uncertainty estimation, which strengthens the model’s robustness in noisy or low-quality conditions. During inference, the framework integrates temporal and spatial selection to adaptively minimize variational free energy, ensuring stable and optimal predictions. The main contributions of this work can be summarized as follows:

\begin{itemize}
    \item We propose an active observation strategy that dynamically selects informative temporal frames and spatial regions, effectively addressing the spatio-temporal sparsity problem in micro-gesture recognition.
    \item We introduce an uncertainty-aware augmentation module (UMIX) that quantifies predictive uncertainty and adaptively re-weights training samples, thereby improving model robustness and generalization under noisy or low-sample conditions.
    \item Extensive experiments on the SMG dataset show that UAAI significantly improves recognition accuracy over state-of-the-art baselines using RGB input. Ablation studies further confirm the effectiveness of each core component and its complementary contributions to the overall performance.
\end{itemize}

\section{Related Works}
\label{sec:formatting}

\subsection{Micro-gesture Recognition}

Micro-gesture recognition aims to identify involuntary, low-amplitude body movements that typically last less than 0.5 seconds. Early studies relied on hand-crafted descriptors such as dense optical flow~\cite{wang2013dense} and spatio-temporal interest points~\cite{laptev2008learning}, which were highly sensitive to noise and lacked the ability to represent complex motion patterns. The rise of deep learning brought substantial improvements. Two-stream networks~\cite{simonyan2014two} fused spatial and motion cues from RGB and optical flow, while 3D CNNs such as C3D~\cite{tran2015learning} jointly modeled spatial and temporal dynamics. Sequence models, including LSTM~\cite{donahue2015long} and TCN~\cite{lea2017temporal}, further captured long-range dependencies, and Transformer-based architectures~\cite{vaswani2017attention,bertasius2021space} achieved impressive results through global spatio-temporal attention. Despite these advances, existing methods still process all frames and spatial regions indiscriminately, failing to focus on transient and task-relevant micro-movements. This passive observation limits efficiency and robustness, leaving the challenge of effectively capturing subtle and temporally sparse micro-gesture signals largely unresolved.

In addition, keyframe selection has been studied in long-video understanding and video reasoning. Methods such as Logic-in-Frames~\cite{guo2025logic} and VideoTree~\cite{wang2025videotree} focus on selecting semantically representative frames to support structured reasoning or LLM-based analysis. These approaches typically operate at a coarse temporal scale and optimize semantic consistency. In contrast, micro-gesture recognition requires fine-grained temporal discrimination under noisy and temporally sparse conditions. Our work differs in that temporal selection is explicitly formulated as an uncertainty-minimization problem under the active inference framework. And unlike saliency mechanisms that passively weight features, our method formulates frame selection as uncertainty-minimizing action under a generative belief model. The policy is derived by minimizing Expected Free Energy, enabling belief updating and active decision-making rather than heuristic attention.

\subsection{Active Inference}

Active inference~\cite{friston2010free,friston2017active,pezzulo2024active} is a theoretical framework rooted in the free-energy principle and the Bayesian brain hypothesis. It assumes that intelligent agents minimize variational free energy by actively sampling sensory data to reduce uncertainty about latent states. This framework has been widely applied to model both biological and artificial systems, demonstrating strong adaptability under uncertain conditions. In robotics, active inference has been used for reactive planning and control, where agents minimize prediction error by dynamically adjusting actions~\cite{lanillos2021active,pezzato2023active,sajid2021active}. Recent studies have also explored its use in active perception tasks such as attention-guided exploration~\cite{shiffrin2003modeling,pardyl2024adaglimpse}. However, research on fine-grained behavior analysis within this paradigm remains limited. In this work, we introduce active inference into micro-gesture recognition for the first time, integrating hierarchical generative modeling with spatio-temporal active observation to enable uncertainty-aware perception and joint inference of latent emotional states.

\subsection{Adaptive learning of uncertainty perception}
Uncertainty modeling has become a key direction for enhancing model robustness under noisy or out-of-distribution conditions. Bayesian dropout~\cite{ahmed2023scale} interprets standard dropout as approximate Bayesian inference, allowing neural networks to estimate epistemic uncertainty through stochastic forward passes. Predictive uncertainty estimation~\cite{deng2023uncertainty} introduces a distinction between data (aleatoric) and model (epistemic) uncertainty, providing calibrated confidence measures that better reflect distributional shifts. 

Uncertainty weighting~\cite{landgraf2023u} further extends this principle by adaptively scaling task losses according to predicted uncertainty, enabling models to prioritize more reliable predictions and mitigate overfitting to noisy labels. However, most existing studies treat uncertainty estimation as a post-hoc calibration step, without explicitly incorporating it into the representation learning or optimization process
To overcome this limitation, we propose an uncertainty-aware augmentation module (UMIX). 
Building upon the uncertainty-aware mixture modulation principle in ~\cite{han2022umix}, UMIX further embeds uncertainty into the Expected Free Energy minimization framework, allowing uncertainty to directly influence representation learning and optimization. 
Unlike standard Mixup that applies uniform interpolation, UMIX adaptively adjusts both the mixing ratio and the contribution of each sample according to its estimated uncertainty, thereby improving robustness and convergence stability under noisy and low-sample conditions.

\section{Problem Formulation}

Let $I_{1:T} = \{I_1, \ldots, I_T\}$ denote a sequence of $T$ consecutive frames capturing subtle body movements such as fingertip twitches or small posture changes. Each sequence corresponds to a ground-truth micro-gesture label $y \in \{1, 2, \ldots, K\}$, where $K$ is the total number of gesture categories. At each time step $t$, a deep encoder $f_\theta$ extracts a compact feature representation from the raw input frame $I_t$, 
yielding a latent embedding $z_t = f_\theta(I_t)$. The encoder $f_\theta$ is implemented as a convolutional neural network that maps each frame to its latent embedding $z_t$.

We formulate the micro-gesture recognition task as a \textbf{state inference problem}. The model aims to estimate a latent posterior distribution $q(s|o)$ that approximates the true posterior $p(s|o)$, where $s$ represents the underlying gesture class. The inference process is optimized by minimizing the \textbf{variational free energy (VFE)}, denoted by $\mathcal{F}$, which can be decomposed into two terms: an \textit{accuracy term} and a \textit{complexity term}. Here, the observation variable $o$ denotes the category-level prediction inferred from the visual embedding $z$. The accuracy term corresponds to the cross-entropy loss that encourages the model to fit the training data accurately, while the complexity term acts as a regularization constraint that measures the Kullback--Leibler divergence between the inferred posterior $q(s|o)$ and the prior $p(s|o)$:\begin{equation}
\mathcal{F} = \mathbb{E}_{q(s|o)}[-\log p(o|s)] + \text{KL}[q(s|o) \| p(s|o)].
\end{equation}
The first term ensures that the inferred latent states explain the observed data effectively, and the second term penalizes deviations of $q(s|o)$ from the prior distribution $p(s)$. This trade-off constrains the latent space to remain compact and disentangled, which helps prevent overfitting and improves generalization. In the following sections, we describe how \textbf{active observation} and \textbf{uncertainty-aware augmentation} are incorporated into this framework to optimize inference efficiency and robustness.


\section{Methodology}

\subsection{Framework Overview}

We propose an \textbf{Uncertainty-Aware Active Inference (UAAI)} framework for micro-gesture recognition. 
The core idea is to jointly optimize the model’s \textit{perception} (learning) and \textit{action} (observation selection) 
under a unified principle of variational free energy (VFE) minimization. 

\begin{figure*}[!t]
    \centering
    \includegraphics[width=1\linewidth]{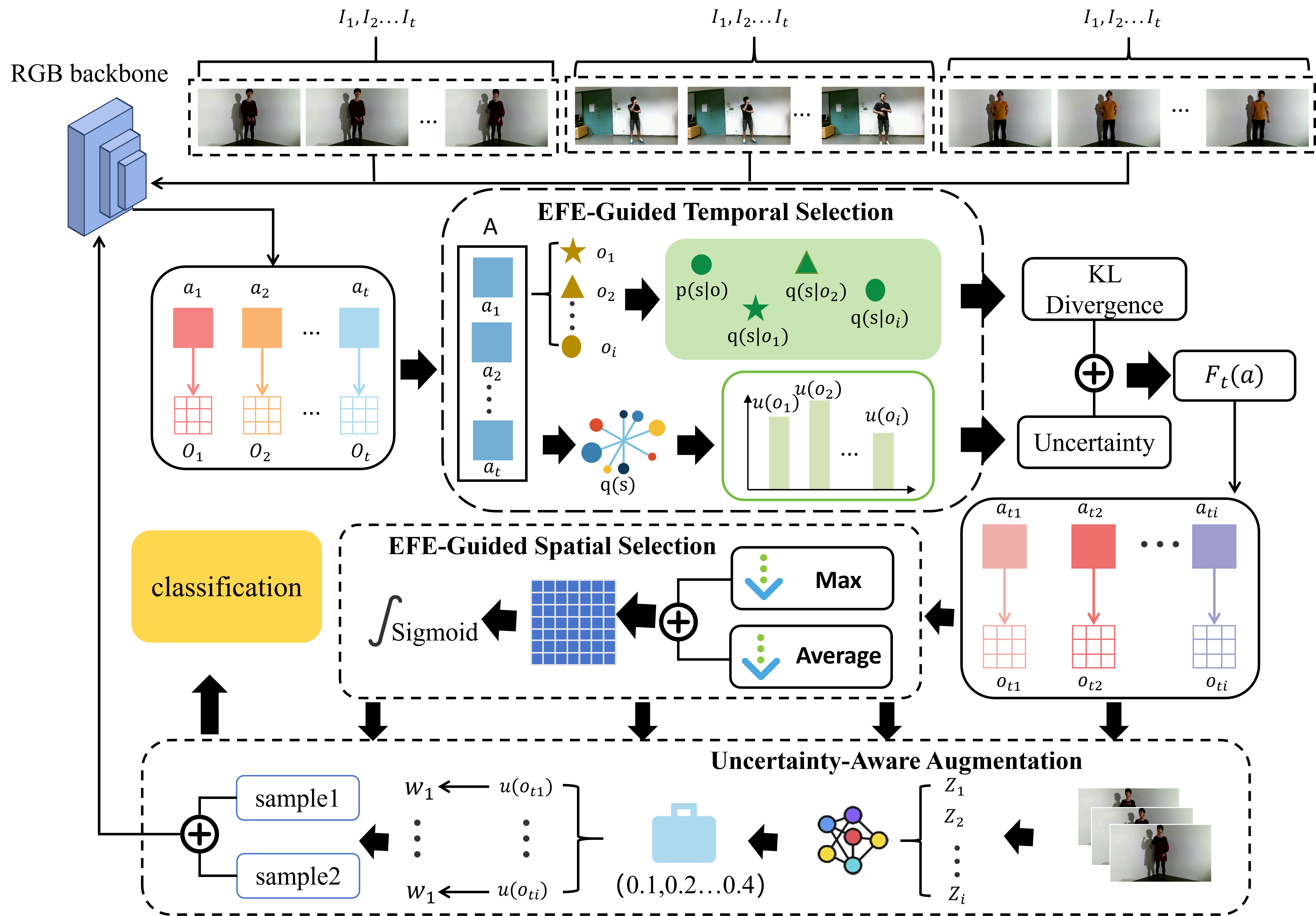}
    \caption{Overall Framework. The framework enhances micro-gesture recognition performance through EFE-based temporal and spatial selection and uncertainty-aware augmentation under the active inference mechanism.}
    \vspace{-0.4cm}
    \label{fig: Overall Framework}
\end{figure*}

We first formalize the micro-gesture recognition task and then instantiate active inference into a two-stage observation strategy 
that addresses the spatiotemporal sparsity challenge inherent to micro-gestures. 
Specifically, we introduce an \textbf{EFE-guided temporal selection module} to capture the most informative keyframes over time dynamically, 
and a \textbf{spatial attention module} to focus on the most discriminative local regions in space. 
Finally, we propose an \textbf{uncertainty-aware augmentation strategy (UMIX)} that leverages the model’s epistemic uncertainty to guide training. 
All modules are jointly optimized from the perspective of minimizing the variational free energy, leading to improved robustness and generalization, especially under noisy or low-sample conditions.The overall
architecture is illustrated in \Cref{fig: Overall Framework}.

\subsection{EFE-Guided Temporal Selection}

Following the active inference paradigm, we formulate the frame selection process in micro-gesture recognition 
as a \textit{Partially Observable Markov Decision Process (POMDP)}. 
The agent aims to choose an action $a_t$ that minimizes the \textit{Expected Free Energy (EFE)} $\mathbb{G}_t$, defined as:

\begin{equation}
a_t^* = \arg\min_{a_t \in \mathcal{A}} \mathbb{G}_t(a_t),
\end{equation}
where $\mathcal{A}$ denotes the set of possible observation actions. 
$\mathbb{G}_t(a_t)$ represents the expected free energy under action $a_t$, computed as:
\begin{equation}
    \begin{aligned}
    \mathbb{G}_t(a_t) &= \mathbb{E}_{q(o_{t+1}, s_{t+1}|a_t)} \big[ D_{\mathrm{KL}}(q(s_{t+1}|o_{t+1}) \Vert p(s_{t+1}|o_{t+1}))\\
    &- \mathbb{H}[(p(o_{t+1}|s_{t+1}, a_t)) \big],
\end{aligned}
\end{equation}
where the first term measures the divergence between the predicted posterior belief and the target belief (epistemic value), 
and the second term quantifies the expected information gain from observation $o_{t+1}$.
 
We parameterize a generative model $\mu_g$ for the likelihood distribution $p(o|s)$ using a lightweight two-layer MLP classifier $\mu(\cdot)$, whose softmax output over gesture classes defines the observation variable $o_t$. 
For a given action $a_t$, its likelihood matrix $\mathbf{A}_{a_t}$ is then defined as:

\begin{equation}
A_{a_t}(i,j) = p(o=j \mid s=i),
\end{equation}
where $i$ indexes the true class and $j$ the observed category. 

In the absence of prior knowledge, we assume a uniform prior over gesture classes, 
$p(s) = \frac{1}{K}$, and compute the posterior belief $q(s'|o',a_t)$ via Bayesian updating. 
The EFE minimization process then selects frames that are expected to most reduce uncertainty about the hidden state $s$. 
Thus, temporal keyframes are selected adaptively to focus on the most informative segments for downstream recognition.

\subsection{EFE-Guided Spatial Selection}








After selecting temporal keyframes, we further optimize spatial perception 
under the same principle of Expected Free Energy (EFE) minimization. 
Since the global EFE in Eq.(3) depends on the observation likelihood, 
and spatial features contribute unequally to this likelihood, 
we decompose the EFE over spatial locations:

\begin{equation}
G_t \approx \sum_i G_{t,i},
\end{equation}

where $G_{t,i}$ denotes the local EFE contribution induced by spatial feature 
$F_i$ at location $i$. To minimize the overall EFE, the model should 
assign larger weights to spatial regions that reduce predictive uncertainty.

We therefore introduce a learnable spatial weighting mask 
$\mathbf{M} \in \mathbb{R}^{H \times W}$, yielding a reweighted feature:

\begin{equation}
\mathbf{F}' = \mathbf{M} \odot \mathbf{F},
\end{equation}

which induces a corresponding weighted EFE:

\begin{equation}
G_t^{spatial} \approx \sum_i M_i \cdot G_{t,i}.
\end{equation}

Minimizing variational free energy thus implicitly optimizes the spatial weights 
$M_i$, encouraging emphasis on informative regions and suppressing irrelevant ones.

In practice, we parameterize $\mathbf{M}$ using a lightweight spatial attention module. 
Given $\mathbf{F} \in \mathbb{R}^{C \times H \times W}$, 
we compute channel-wise average pooling and max pooling:
\begin{equation}
\mathbf{F}_{avg} = \mathrm{AvgPool}_c(\mathbf{F}), \quad 
\mathbf{F}_{max} = \mathrm{MaxPool}_c(\mathbf{F}),
\end{equation}

which are concatenated and passed through a convolution layer 
followed by a sigmoid activation:

\begin{equation}
\mathbf{M} = \sigma(\mathrm{Conv}([\mathbf{F}_{avg}; \mathbf{F}_{max}])).
\end{equation}

This design provides a differentiable approximation of spatial EFE minimization 
while remaining computationally efficient.

\subsection{Uncertainty-Aware Augmentation}

\begin{table*}[!ht]
\centering
\caption{Comparison with state-of-the-art methods on the SMG dataset in different modalities. * denotes our framework.}
\label{tab:comparison}

\begin{minipage}{\textwidth}
\centering
\begin{tabular}{lccccccc}
\hline
Method & GCN-NAS & MS-G3D & MSTCN-VAE & C3D & TSM & TRN & UAAI* \\ \hline
Modality & Skeleton & Skeleton & RGB & RGB & RGB & RGB & RGB \\
Accuracy (\%) & 58.85 & 64.75 & 42.59 & 45.90 & 58.69 & 59.51 & \textbf{63.47} \\ \hline
\end{tabular}
\end{minipage}

\vspace{0.3cm} 

\begin{minipage}{\textwidth}
\centering
\begin{tabular}{lcccccccc}
\hline
Method       & ST-GCN & 2S-GCN & Shift-GCN  & MA-Net & TSN & Video Mamba & UAAI* \\ \hline
Modality     & Skeleton & Skeleton & Skeleton  & RGB & RGB & RGB & RGB \\
Accuracy (\%) & 41.48 & 43.11 & 55.31  & 48.69 & 50.49 & 55.08 & \textbf{63.47} \\ \hline
\end{tabular}
\end{minipage}
\end{table*}

\begin{table*}[!ht]
\centering
\caption{Module Ablation Experiments}
\label{tab: ablation}
\scalebox{1.1}{
\begin{tabular}{c|ccc|c}
\hline
\textbf{Method}     & \textbf{Uncertainty-aware} & \textbf{Temporal Selection} &\textbf{Spatial Selection} & \textbf{Accuracy (\%)} \\ \hline
Baseline            & \textit{No}                & \textit{No}      & \textit{No}           & 50.49                 \\
+Uncertainty-aware  & \textit{Yes}               & \textit{No}      & \textit{No}           & 57.54                 \\
+Temporal Selection & \textit{No}                & \textit{Yes}      & \textit{No}          & 56.40                 \\
+Spatial Selection & \textit{No}                & \textit{No}       & \textit{Yes}         & 55.40                 \\
UAAI                & \textit{Yes}               & \textit{Yes}         & \textit{Yes}       & \textbf{63.47}  
\\ \hline
\end{tabular}
}
\end{table*}

Minimizing free energy requires an accurate estimation of observation uncertainty. 
To this end, we introduce a Monte Carlo Dropout-based uncertainty estimation module to quantify epistemic uncertainty for each training sample. 
For each sample $(I, y)$, after temporal and spatial selection, we perform $T$ stochastic forward passes with dropout active, yielding a set of predicted distributions $\{\hat{p}_t(y|I)\}_{t=1}^T$. 
The uncertainty score $u(I)$ is then defined as the maximum variance across classes:
\begin{equation}
u(I) = \max_k \mathrm{Var}_t[\hat{p}_t(y=k|I)].
\end{equation}
A higher $u(I)$ indicates unstable predictions, suggesting that the sample is noisy or hard to classify. 

We then assign a weight $w_i$ to each sample according to its uncertainty:
\begin{equation}
w_i = \exp(-\alpha \cdot u(I_i)) + \beta,
\end{equation}
where $\alpha$ and $\beta$ are hyperparameters controlling sensitivity.

Next, we perform soft sample mixing. 
Given two randomly selected samples $(x_i, y_i)$ and $(x_j, y_j)$, 
a mixing coefficient $\lambda \sim \mathrm{Beta}(\alpha_{mix}, \alpha_{mix})$ is drawn, and new augmented samples are generated as:
\begin{equation}
\tilde{x} = \lambda x_i + (1-\lambda)x_j, \quad 
\tilde{y} = \lambda y_i + (1-\lambda)y_j.
\end{equation}

The final training loss is then expressed as:

\begin{equation}
\begin{aligned}
    \mathcal{L} &= \mathbb{E}_{(x_i,y_i), (x_j, y_j)}[w_i\lambda \mathcal{L}_{ce}(\theta, \tilde{x}, y_i) \\ 
    &+ w_j(1-\lambda)(\theta, \tilde{x}, y_j)],
\end{aligned}
\end{equation}
where $ \mathcal{L}_{ce}(\cdot)$ denotes the cross-entropy loss. By weighting and mixing based on uncertainty, UMIX imposes a data-driven smooth regularization effect, acting as a dynamic implicit regularizer that alleviates overfitting and improves generalization.

In summary, our UAAI framework integrates temporal and spatial active inference with uncertainty-aware augmentation, achieving an elegant balance between efficiency, robustness, and interpretability.
\section{Experiments}

\subsection{Experimental Settings}

\textbf{Dataset.} We conduct all experiments on the SMG dataset~\cite{chen2023smg}, a large-scale benchmark designed for spontaneous micro-gesture recognition and hidden emotional stress analysis. The dataset was collected at the University of Oulu using two Kinect V2 sensors and includes recordings of forty participants from diverse cultural backgrounds (average age 25.8). It contains 488 minutes and 821,000 frames of data across four modalities: RGB, depth, contour, and skeleton. To induce emotional variation, two psychological conditions were designed: \textit{neutral expressive storytelling (NES)} representing low stress and \textit{stressful expressive storytelling (SES)} representing high stress. Each one-minute video is annotated with emotion states, and each frame is labeled with 16 micro-gesture classes plus one non-gesture class, resulting in 3,712 micro-gesture segments with an average length of 51.3 frames. Statistical analysis shows that micro-gesture frequency is significantly higher under stress (38.58 vs. 24.50 per participant, $p < 0.0001$), while strong inter-individual differences remain ($r = 0.456$). The dataset provides subject-independent and semi-subject-independent splits for evaluation.We follow the official data split, which contains 2,452 training samples, 637 validation samples, and 610 test samples across 17 gesture categories.

\textbf{Baselines.} We compared UAAI with two kinds of baselines: RGB-based and skeleton-based methods. RGB-based methods include two-dimensional CNN methods, such as TSM, modeling time dynamics from two-dimensional features, and MSTCN-VAE~\cite{yuan2023mstcn}, MA-Net~\cite{cui2023ma}, and Video-Mamba~\cite{li2024videomamba}. The baseline based on the skeleton is the graph convolution network (GCN) model, including ST-GCN~\cite{yan2018spatial}, 2S-GCN~\cite{guo2024skeleton}, and Shift-GCN~\cite{cheng2020skeleton}. Joint-level reasoning is performed on the skeleton data.

\begin{figure*}[!t]
    \centering
    \includegraphics[width=0.9\linewidth]{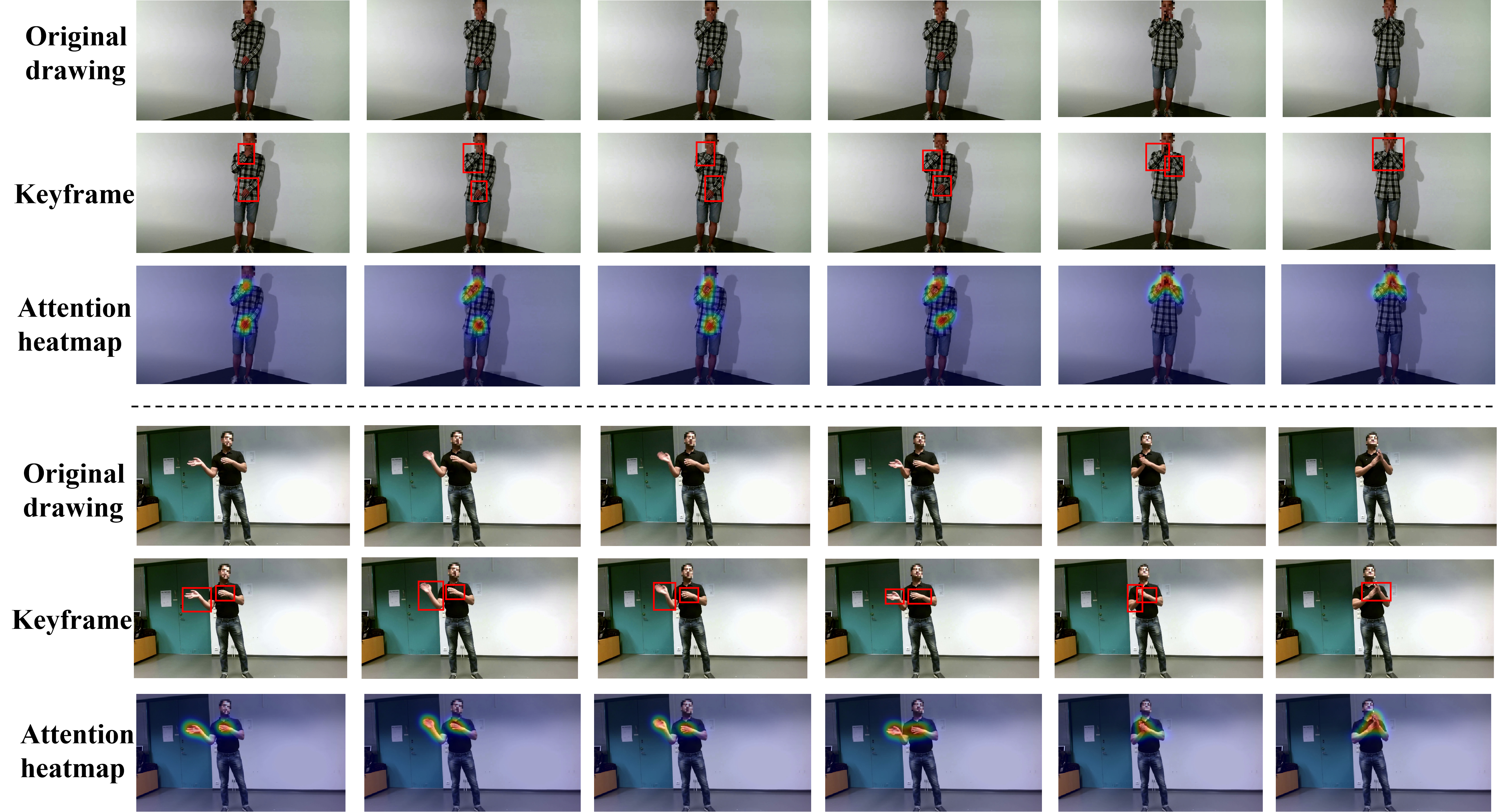}
    \caption{Active Observation Visualization}
    \label{fig: visuallization}
    \vspace{-0.4cm}
\end{figure*}

\textbf{Implementation Details.} Both subject-independent and semi-subject-independent evaluation protocols are adopted to ensure cross-subject generalization. Input frames are resized to $224\times224$ and processed by a ResNet-50 backbone pre-trained on ImageNet. The models are optimized using the Adam optimizer with a batch size of 32. The initial learning rate is selected by cross-validation within $[1\times10^{-4}, 5\times10^{-4}]$. Dropout with a rate of 0.3 is applied during training and is also used for Monte Carlo sampling ($T=5$) in uncertainty estimation. UMIX augmentation is employed with sample-level mixing weights defined as $(1-u)$, where $u$ denotes the uncertainty score, and the mixing coefficient $\lambda$ is drawn from $\text{Beta}(\alpha,\alpha)$ with $\alpha=0.4$.

\textbf{Evaluation Metric.} Classification accuracy (\%) is reported as the mean of three independent runs. We also report the average number of observed frames per sequence to evaluate computational efficiency.

\subsection{Comparative Experiments}

~\Cref{tab:comparison} compares UAAI with existing micro-gesture recognition methods. UAAI achieves the best performance among all RGB-based baselines. This improvement arises from the framework’s ability to select key frames and regions that contain discriminative gesture cues while minimizing redundant information through free-energy optimization. By jointly modeling temporal and spatial dynamics under active inference, UAAI significantly enhances robustness to noise and improves learning efficiency. Although skeleton-based methods such as MS-G3D achieve slightly higher absolute accuracy, UAAI narrows the gap to within 1.28 percentage points using only RGB data, which is easier to acquire in real-world scenarios.Beyond micro-gesture baselines, we further compare with representative keyframe selection strategies designed for long-video understanding as shown in ~\Cref{tab:Comparison of frame selection methods}.

\begin{table}[h]
\centering
\setlength{\abovecaptionskip}{0.cm}
\caption{Comparison of frame selection methods}
\label{tab:Comparison of frame selection methods}
\scalebox{0.8}{\begin{tabular}{|c|c|c|}
\hline
Frame Selection & Dataset & Accuracy (\%) \\ \hline
Logic-in-Frames~\cite{guo2025logic} & SMG     & 61.31         \\ \hline
Video Tree~\cite{wang2025videotree}      & SMG     & 59.51         \\ \hline
\textbf{Ours}   & SMG     & \textbf{63.47}         \\ \hline
\end{tabular}}
\end{table}


\begin{figure*}[!t]
    \centering
    \subfloat[20 epochs]{\includegraphics[width=0.18\linewidth]{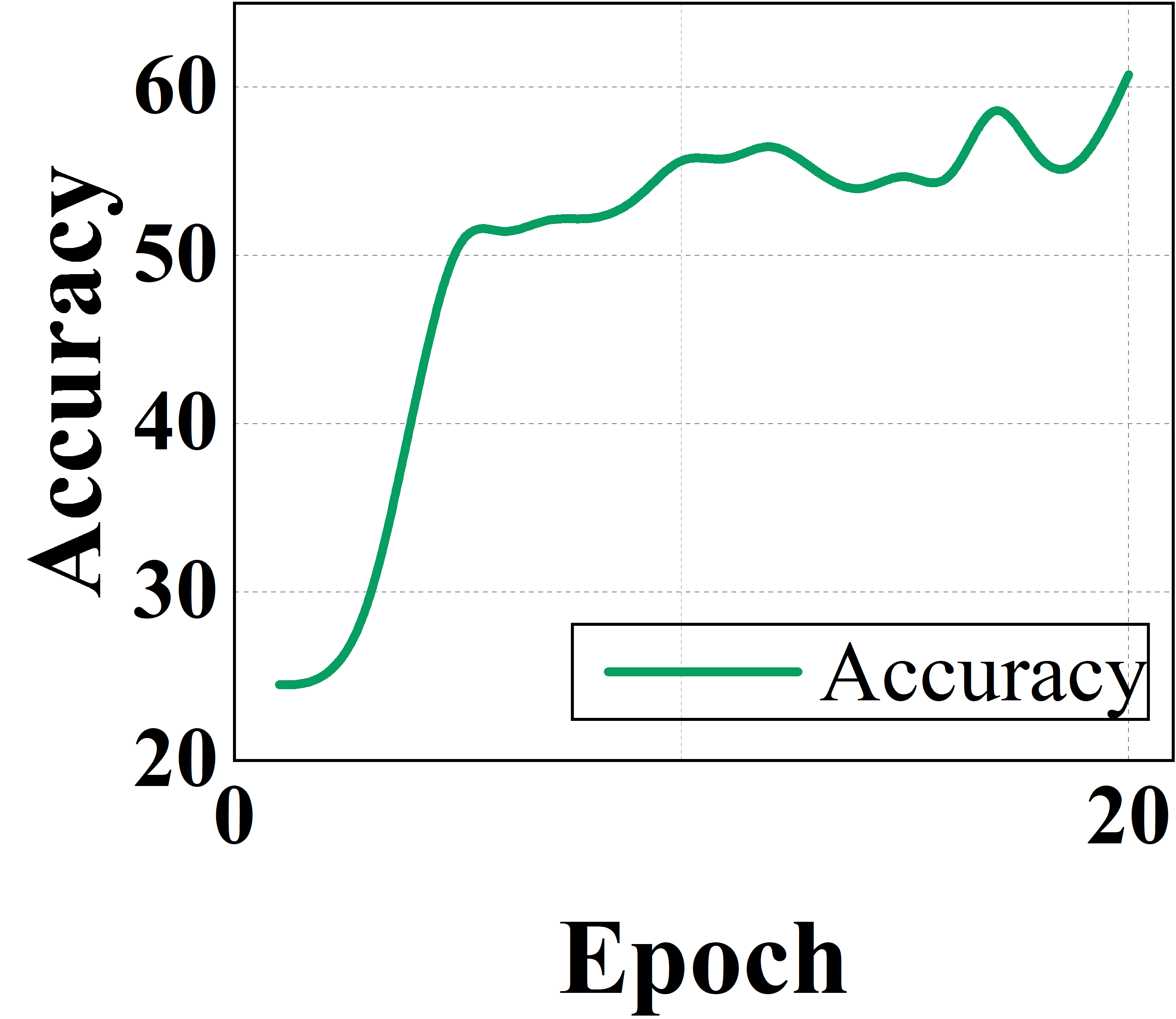}}
    \hfill
    \subfloat[30 epochs]{\includegraphics[width=0.18\linewidth]{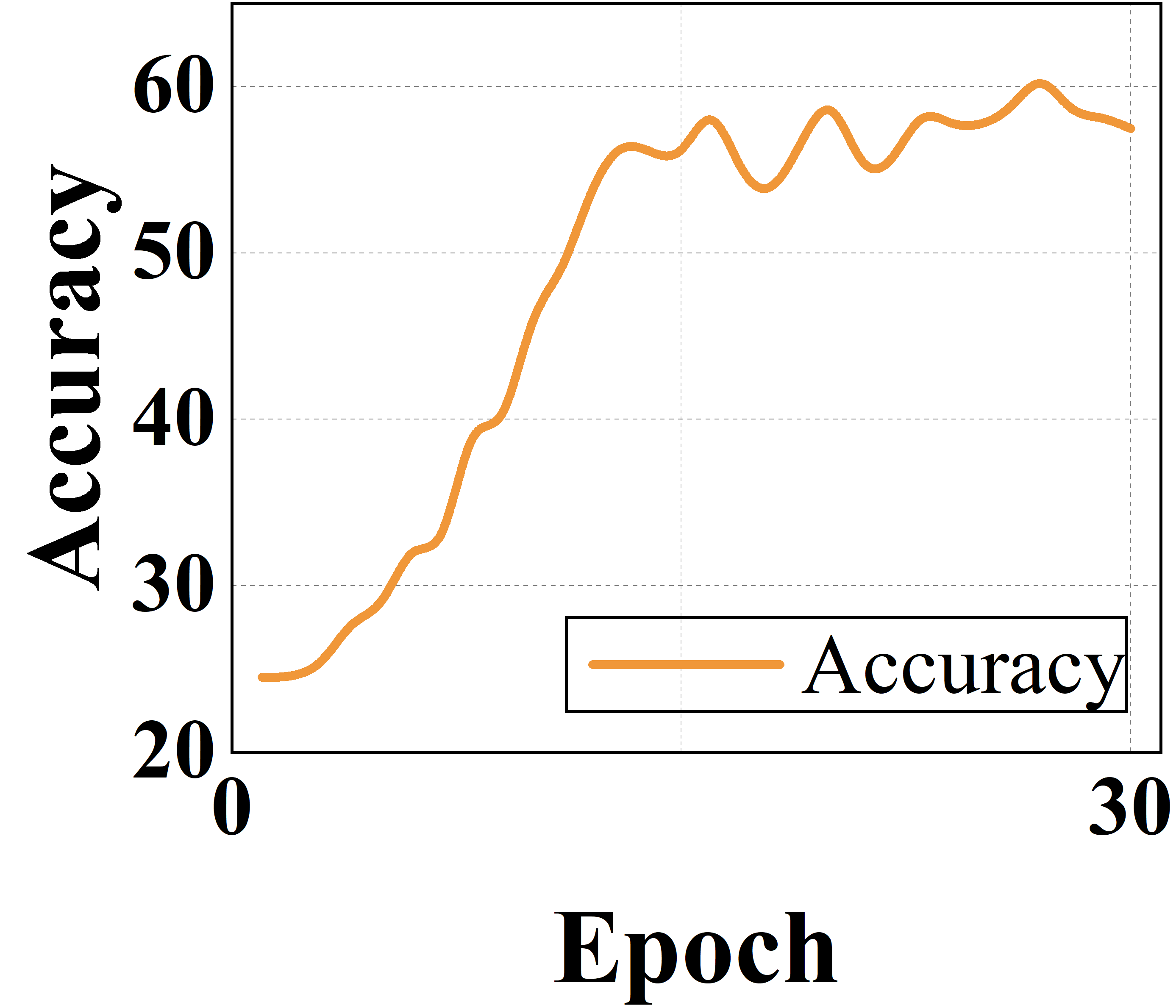}}
    \hfill
    \subfloat[40 epochs]{\includegraphics[width=0.18\linewidth]{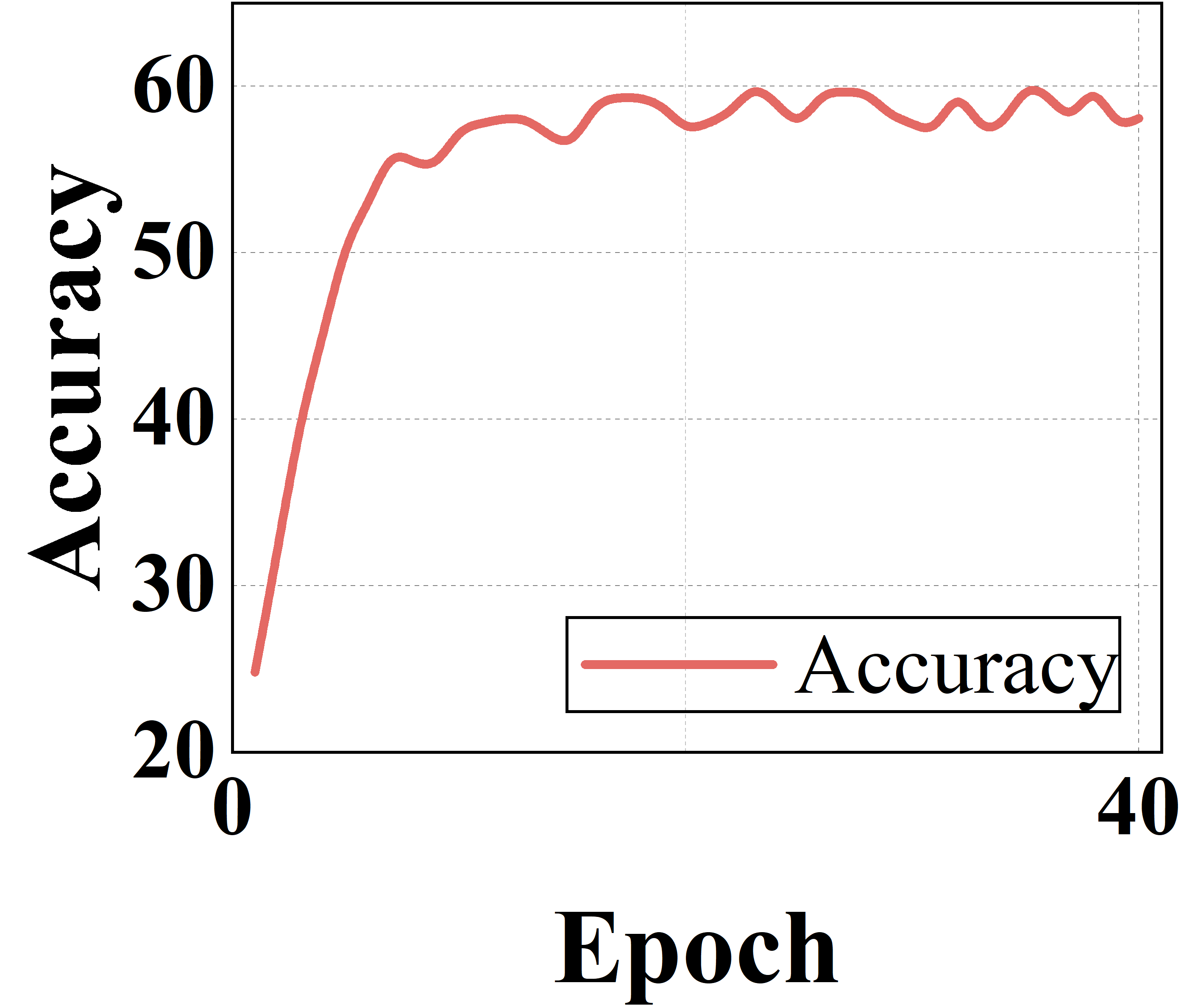}}
    \hfill
    \subfloat[50 epochs]{\includegraphics[width=0.18\linewidth]{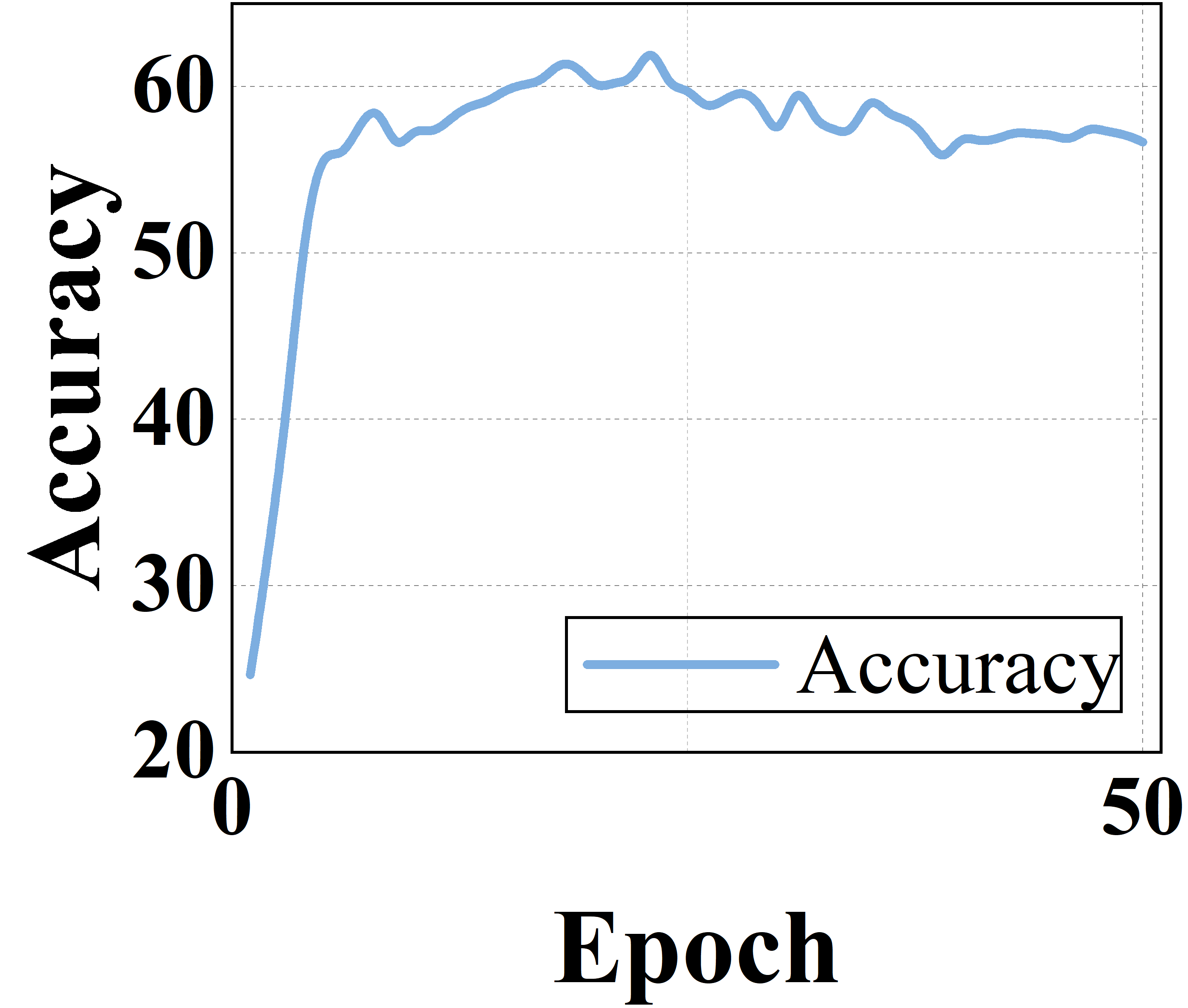}}
    \hfill
    \subfloat[60 epochs]{\includegraphics[width=0.18\linewidth]{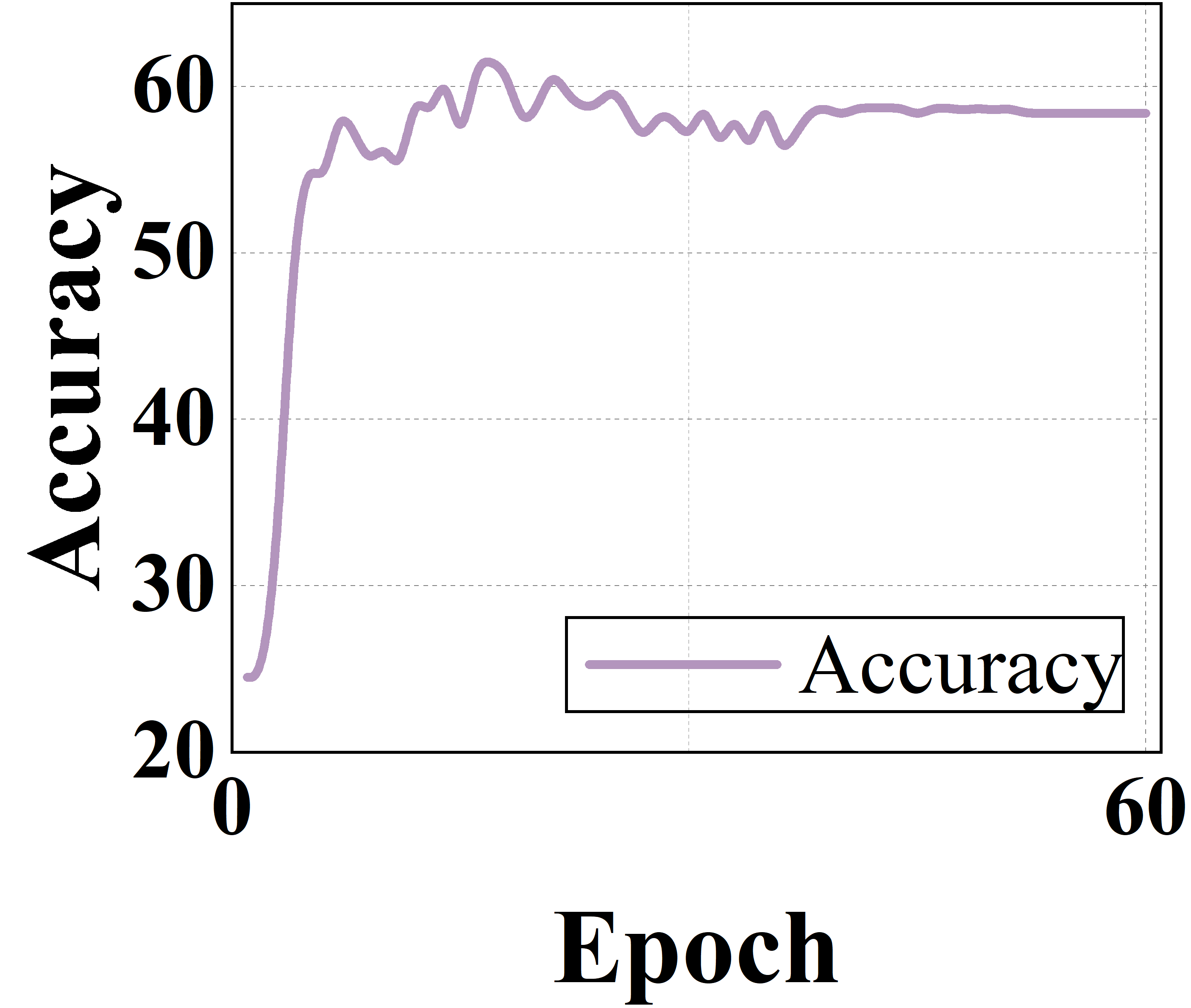}}

    \subfloat[20 epochs]{\includegraphics[width=0.18\linewidth]{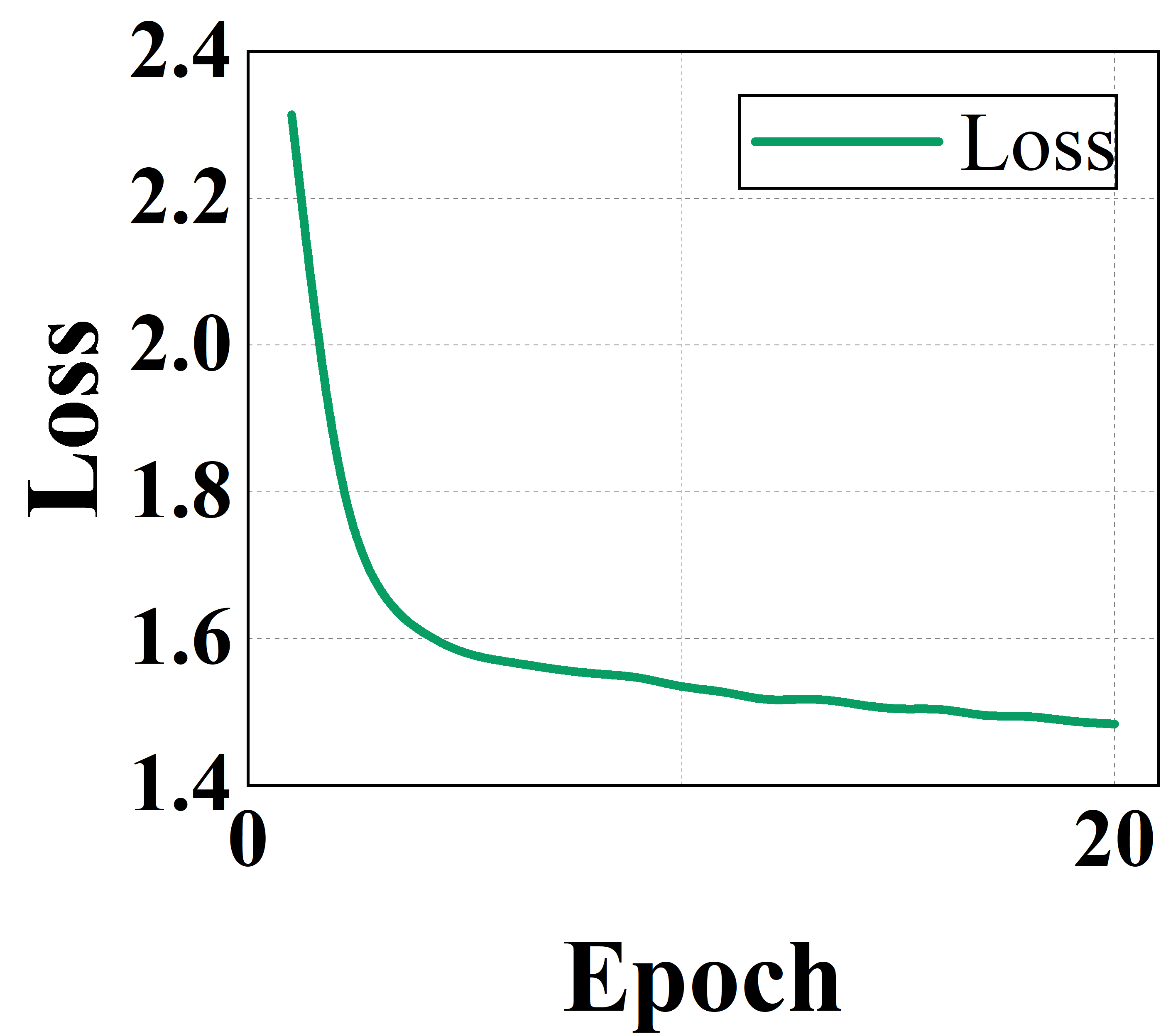}}
    \hfill
    \subfloat[30 epochs]{\includegraphics[width=0.18\linewidth]{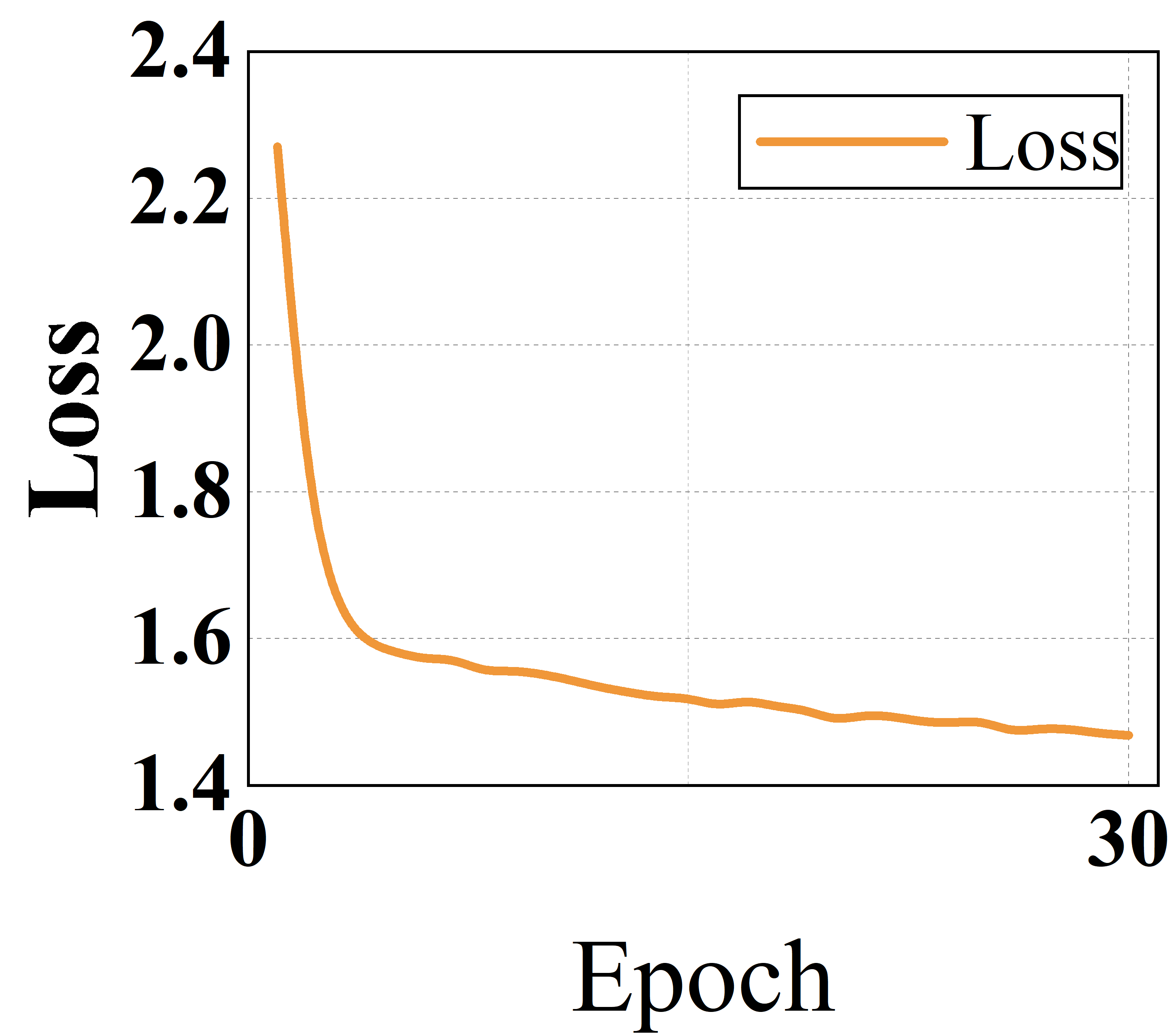}}
    \hfill
    \subfloat[40 epochs]{\includegraphics[width=0.18\linewidth]{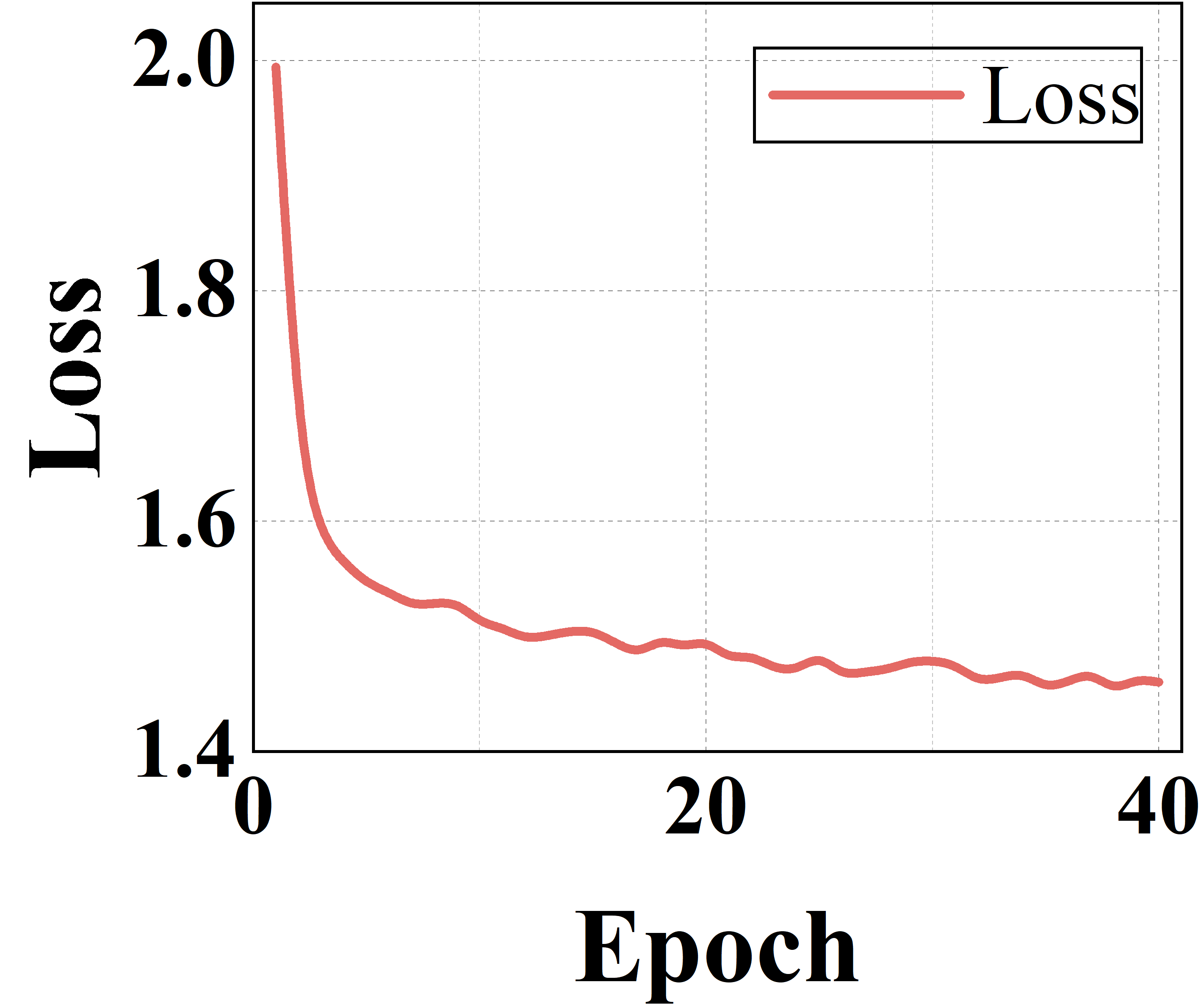}}
    \hfill
    \subfloat[50 epochs]{\includegraphics[width=0.18\linewidth]{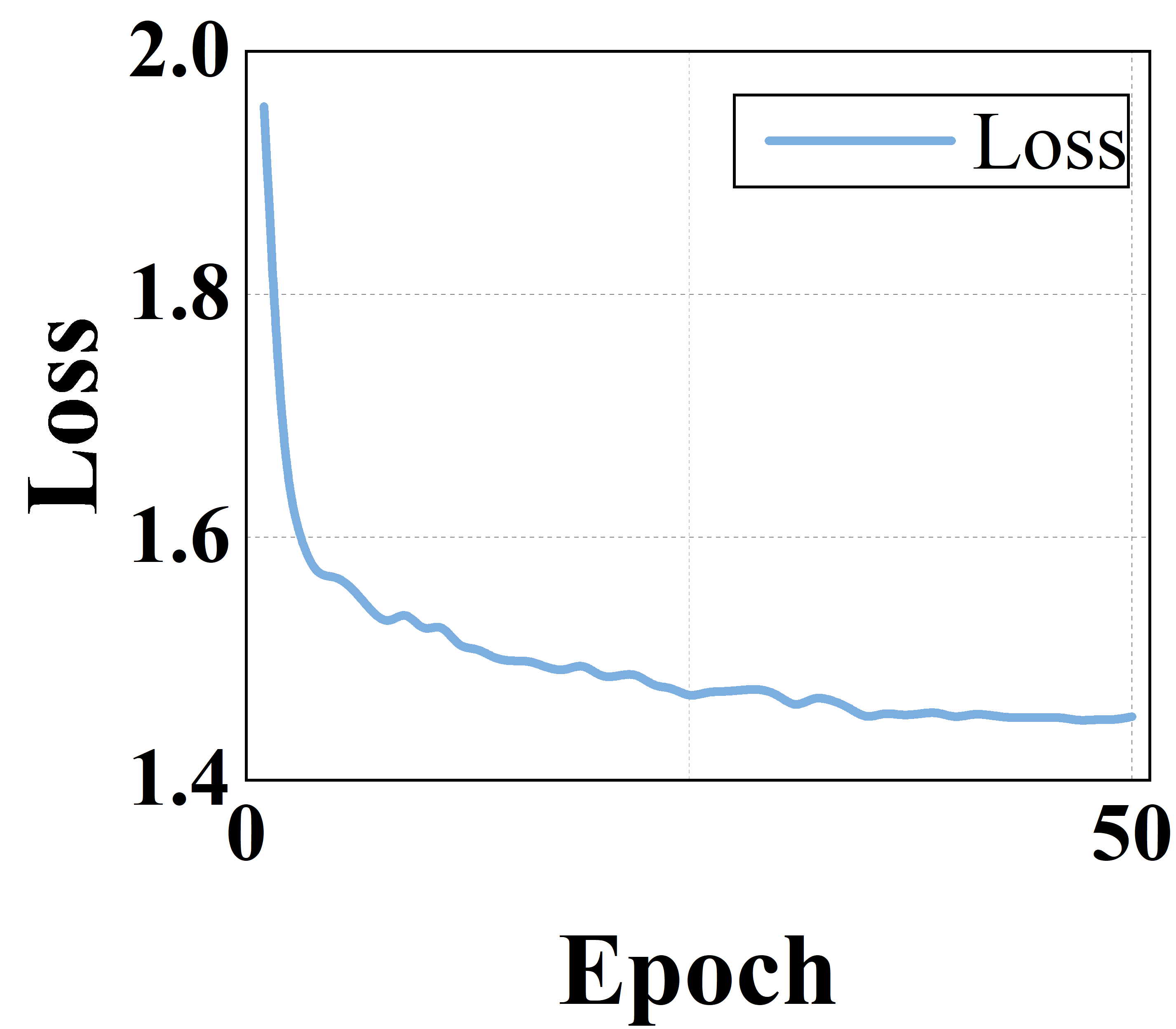}}
    \hfill
    \subfloat[60 epochs]{\includegraphics[width=0.18\linewidth]{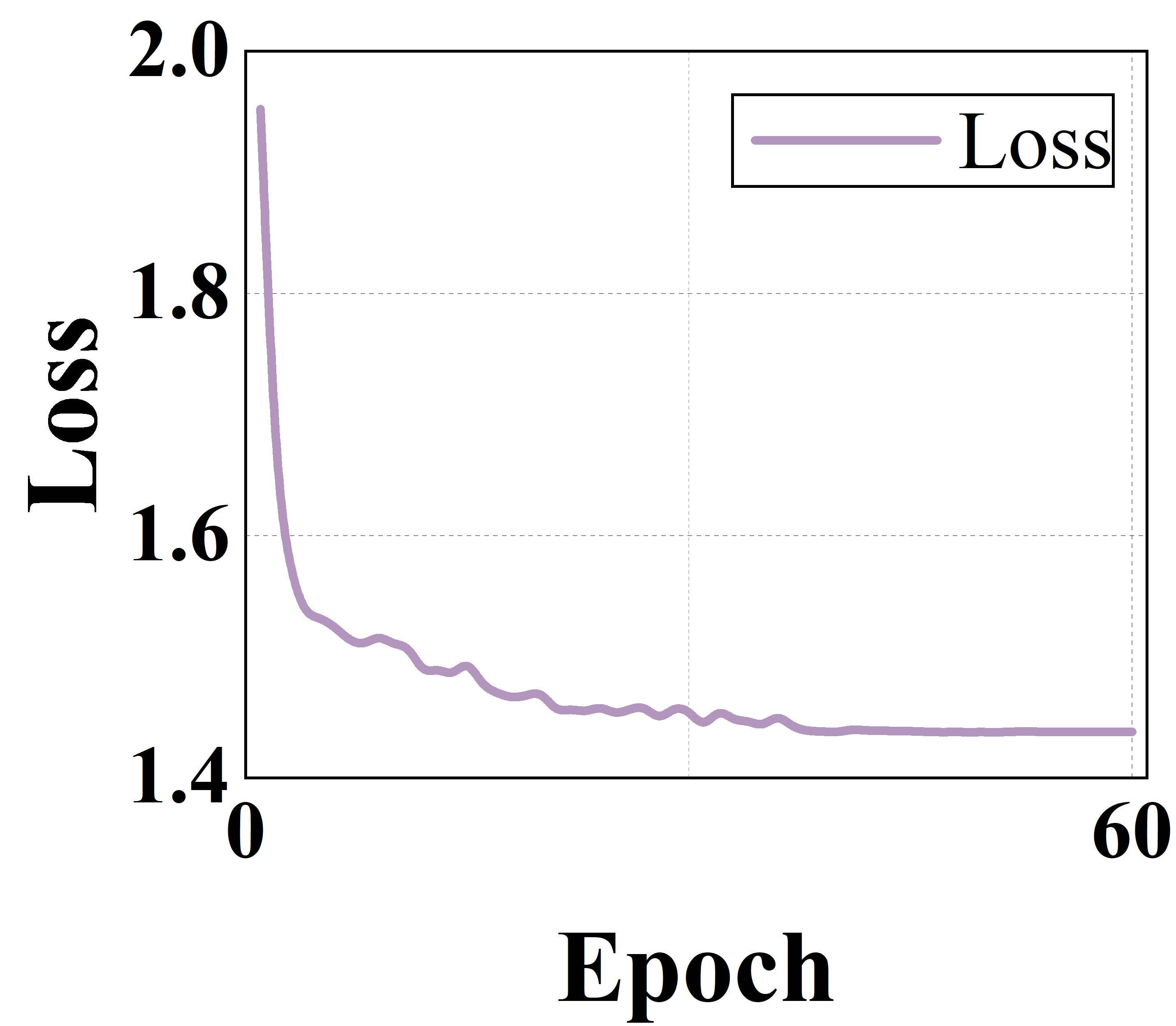}}
    \caption{Convergence curves of UAAI under different training epochs. The upper row shows accuracy curves, and the lower row shows corresponding loss curves. The model converges stably after around 40 epochs.}
    \label{fig:epoch}
    \vspace{-0.4cm}
\end{figure*}

\subsection{Convergence Analysis}
To examine the convergence behavior and stability of the proposed UAAI framework, we trained the model for different numbers of epochs (20, 30, 40, 50, and 60) and plotted the corresponding accuracy and loss curves, as shown in \Cref{fig:epoch}.
The experimental results reveal that both training and validation losses decrease rapidly during the initial stages and gradually stabilize after approximately 40 epochs. Concurrently, the accuracy curves show consistent improvement, eventually saturating around 60\% accuracy. This convergence pattern demonstrates the model's efficient learning capability and effective resistance to overfitting. The findings suggest that training for 40-50 epochs achieves an optimal balance between convergence efficiency and performance stability, providing practical guidance for deployment in real-world applications.

\subsection{Ablation Studies}

\begin{figure}[htbp] 
    \centering
    \begin{minipage}[b]{0.48\columnwidth} 
        \includegraphics[width=0.93\linewidth]{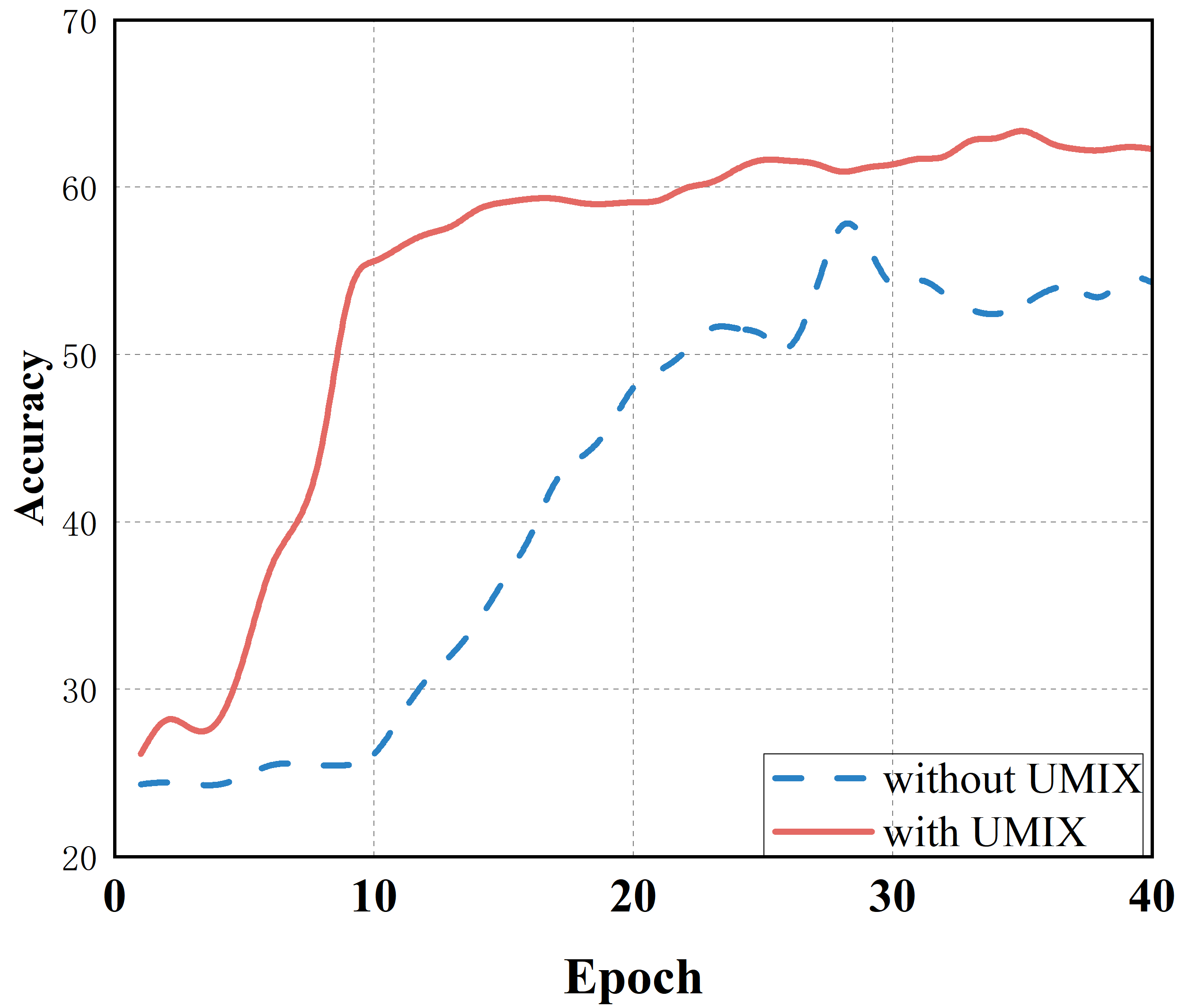}
        \label{fig:umix-acc}
    \end{minipage}
    \hfill 
    \begin{minipage}[b]{0.48\columnwidth} 
        \includegraphics[width=0.93\linewidth]{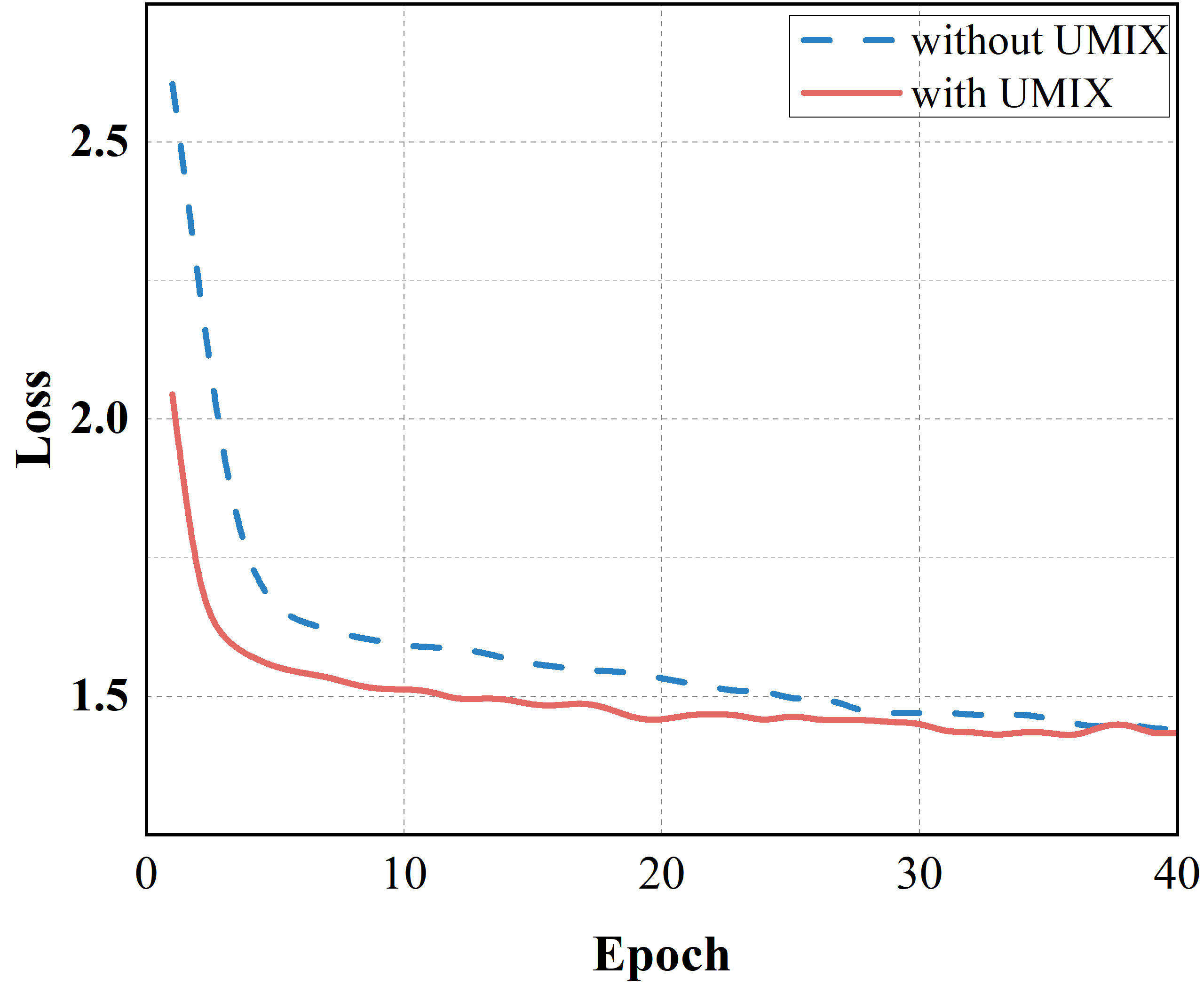}
        \label{fig:umix-loss} 
    \end{minipage}
    \caption{Convergence curves with and without the UMIX.}
    \label{fig: umix} 
    \vspace{-0.4cm }
\end{figure}

To systematically assess the contribution of each core component, we perform ablation experiments by enabling or disabling specific modules within UAAI. As shown in \Cref{tab: ablation}, the baseline model without any components achieves 50.49\% accuracy. When the \textbf{uncertainty-aware augmentation (UMIX)} module is enabled, accuracy increases to 57.54\%, confirming that uncertainty-guided reweighting improves robustness to noisy samples. 
As further shown in \Cref{fig: umix}, the model with UMIX reaches high-performance at earlier epochs and exhibits faster loss reduction with improved training stability compared to the baseline. This demonstrates that UMIX improves robustness without slowing down convergence.
When only the \textbf{EFE-guided temporal selection} is applied, accuracy reaches 56.40\%, demonstrating the effectiveness of active observation for capturing key temporal cues. 
When only the EFE-guided spatial selection is applied, the accuracy reaches 55.40\%, demonstrating the effectiveness of active observation for capturing key spatial cues.When all modules are jointly applied, the full UAAI model achieves 63.47\% accuracy, outperforming all RGB-based baselines and approaching skeleton-based state-of-the-art models. These results validate the complementary synergy between uncertainty modeling and free-energy-based temporal selection.

\subsection{Further Remarks}
To investigate the sensitivity of key hyperparameters in the model and determine their optimal configurations, we conducted a series of ablation experiments on the SMG dataset. Using the final classification accuracy (Accuracy) as the primary evaluation metric, we focused on assessing the effects of the number of Monte Carlo
sampling. The goal is to provide empirical support for our parameter selection and to validate the effectiveness of the proposed model design.

\begin{figure}[tbp] 
    \centering 
    \begin{minipage}[b]{0.48\columnwidth} 
        \includegraphics[width=\linewidth]{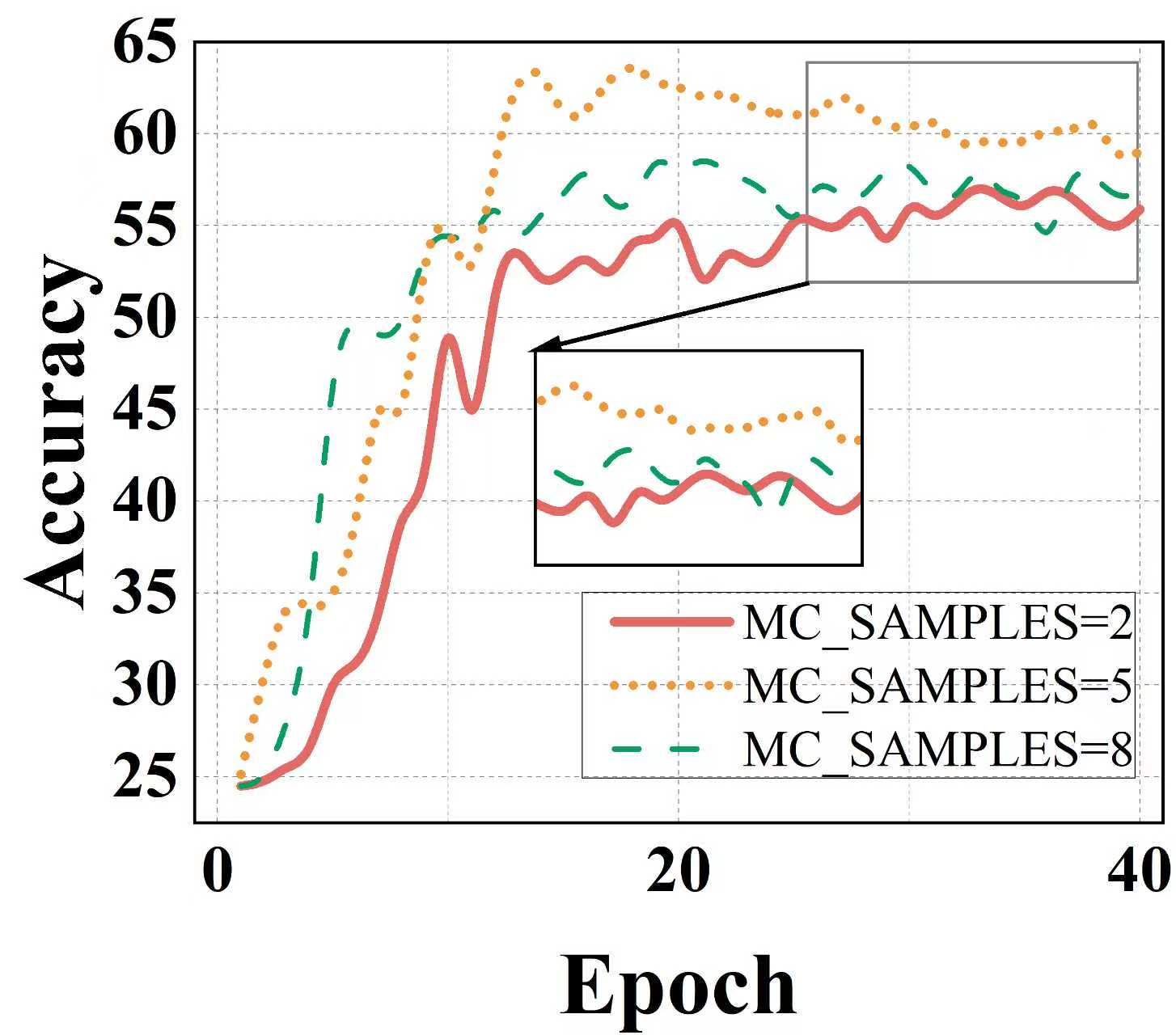}
        \subcaption{Accuracy Curve} 
        \label{fig:mcaccuracy_curve}
    \end{minipage}
    \hfill 
    \begin{minipage}[b]{0.48\columnwidth} 
        \includegraphics[width=\linewidth]{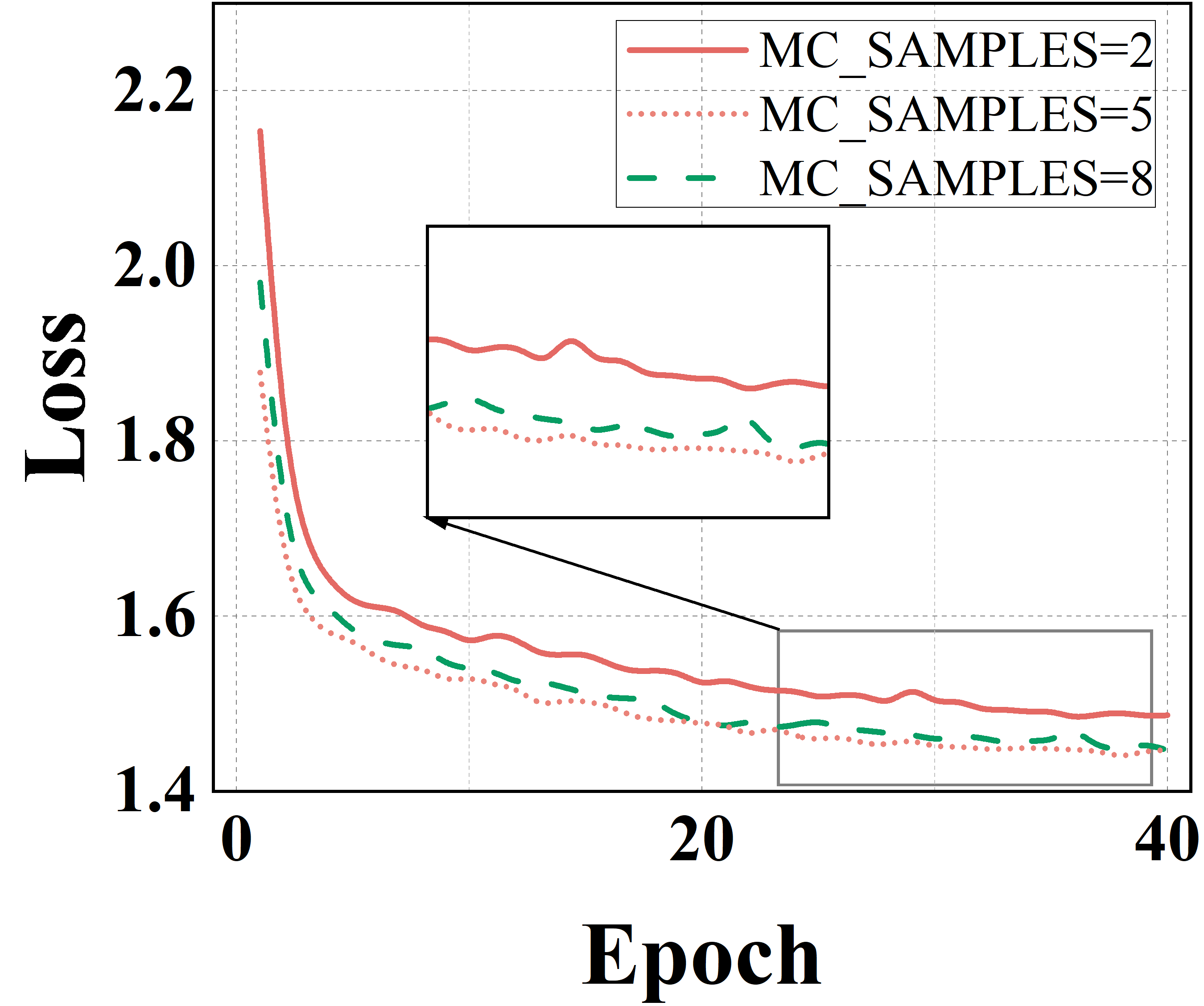}
        \subcaption{Loss Curve} 
        \label{fig:mcloss_curve} 
    \end{minipage}
    \caption{Accuracy and loss Curves under different Monte Carlo sampling numbers M}
    \label{fig: MC_acc_loss_curves}
    \vspace{-0.4cm }
\end{figure}

To analyze the influence of the Monte Carlo sampling number M on uncertainty estimation and training behavior, we evaluate three settings (M = 2, 5, 8) and report the validation accuracy and loss curves in \Cref{fig: MC_acc_loss_curves}. The results show that M = 5 achieves the best overall performance, enabling faster and more stable convergence. Smaller values (M = 2) lead to unstable fluctuations due to higher estimation variance, while larger values (M = 8) provide negligible performance gains.

We further quantify the associated computational overhead in ~\Cref{tab:Training cost}. As expected, the training wall-clock time increases approximately linearly with M, whereas GPU utilization remains relatively stable. These results indicate that although uncertainty estimation requires multiple stochastic forward passes, the additional training cost is moderate. Therefore, M = 5 offers a favorable trade-off between estimation quality and computational efficiency and is adopted as the default configuration.

\begin{table}[h!]
\centering
\caption{Training cost under different M}
\label{tab:Training cost}
\scalebox{0.68}{
\begin{tabular}{|c|c|c|c|}
\hline
M  & Training wall-clock time(s) & Throughput (samples/s) & GPU utilization (\%) \\ \hline
2  & $\sim$152                   & $\sim$645              & $\sim$9.2          \\
5  & $\sim$204                   & $\sim$523              & $\sim$9.7        \\
8 & $\sim$255                   & $\sim$392              & $\sim$10.1       \\ \hline
\end{tabular}
}
\vspace{-0.5cm}
\end{table}

\subsection{Qualitative Analysis}

To better understand the behavior of our model, we visualize spatio-temporal attention maps generated by UAAI on test samples from the SMG dataset, as shown in \Cref{fig: visuallization}. The visualizations show that the model concentrates on body parts closely related to micro-gesture expression, such as fingers, hands, and the upper torso, while suppressing irrelevant background regions like walls and shadows. This focused attention confirms that the spatial selection mechanism effectively learns to ignore redundant information and enhance discriminative features. The resulting interpretability and robustness make UAAI suitable for deployment in complex real-world environments.

\section{Conclusion}


This paper presents UAAI, a novel active inference framework for micro-gesture recognition that integrates Expected Free Energy (EFE)-guided temporal and spatial sampling with uncertainty-aware augmentation (UMIX) into a unified variational free-energy minimization objective. By actively selecting discriminative spatio-temporal cues and adaptively reweighting samples based on epistemic uncertainty, our approach effectively addresses the inherent challenges of micro-gesture signals, including temporal sparsity, spatial subtlety, and label noise. Extensive experiments on the SMG dataset demonstrate that UAAI achieves 63.47\% accuracy, setting a new state-of-the-art among RGB-based methods and significantly narrowing the performance gap with skeleton-based models. Our work provides a robust and interpretable paradigm for fine-grained behavior understanding, offering broad applicability to human-computer interaction and clinical affective computing.
\section{Acknowledgements}
This study is funded in part by the National Key Research and Development Plan: 2019YFB2101900, NSFC (Natural Science Foundation of China): 61602345, 62002263, Key Project of Tianjin Natural Science Foundation: 25JCZDJC00250, TianKai Higher Education Innovation Park Enterprise R\&D Special Project: 23YFZXYC00046. 

\newpage
{
   \small
   \bibliographystyle{ieeenat_fullname}
   \bibliography{main}
}

\end{document}